%% file: main.tex
\documentclass[letterpaper, 10 pt, conference]{ieeeconf}
\IEEEoverridecommandlockouts                              
\usepackage{amsmath,bm,amsfonts,cases}
\usepackage{shortcuts,subfiles,balance}
\usepackage{booktabs,xcolor}
\usepackage{graphicx}
\usepackage{float}
\usepackage{url}
\usepackage{optidef}    
\usepackage{mathrsfs}
\usepackage[hidelinks]{hyperref} 

\usepackage{multirow}
\usepackage{colortbl}
\usepackage{siunitx}    

\let\labelindent\relax
\usepackage{enumitem}

\usepackage{tabularx}
\usepackage[format=plain,font=footnotesize,labelsep=colon]{caption}

\title{\LARGE \bf Chasing Autonomy: Dynamic Retargeting and Control Guided RL for Performant and Controllable Humanoid Running }
\author{Zachary Olkin, William D. Compton, Ryan M. Bena, Aaron D. Ames$^{}$
\thanks{The authors are with the Department of Control and Dynamical Systems and the Department of Mechanical and Civil Engineering at the California Institute of Technology. 
This research is supported by the Technology Innovation Institute (TII).}
}

\begin{document}
\bstctlcite{BSTcontrol}
\maketitle

\begin{abstract}
Humanoid robots have the promise of locomoting like humans, including fast and dynamic running. Recently, reinforcement learning (RL) controllers that can mimic human motions have become popular as they can generate very dynamic behaviors, but they are often restricted to single motion play-back which hinders their deployment in long duration and autonomous locomotion. In this paper, we present a pipeline to dynamically retarget human motions through an optimization routine with hard constraints to generate improved periodic reference libraries from a single human demonstration. We then study the effect of both the reference motion and the reward structure on the reference and commanded velocity tracking, concluding that a goal-conditioned and control-guided reward which tracks dynamically optimized human data results in the best performance. We deploy the policy on hardware, demonstrating its speed and endurance by achieving running speeds of up to 3.3 m/s on a Unitree G1 robot and traversing hundreds of meters in real-world environments. Additionally, to demonstrate the controllability of the locomotion, we use the controller in a full perception and planning autonomy stack for obstacle avoidance while running outdoors.
\end{abstract}

\subfile{sections/introduction}

\subfile{sections/preliminaries}

\subfile{sections/methods}
\subfile{sections/results}

\section{Conclusion}
\label{sec:conclusion}
In this paper we presented a pipeline for using optimization with hard constraints to dynamically retarget human data and then track this with CLF-RL. We studied both the effect of the reference motion and the reward on the reference and velocity command tracking. We concluded that goal conditioned control-guided RL (i.e. CLF-RL) in conjunction with dynamic optimized human data yields the best tracking performance. The resulting controller is applied on a real humanoid robot which achieves controllable and performant running in outdoor real world environments. Further, the running controller is integrated into an autonomy stack for perceptive real-time obstacle avoidance and navigation.

\bibliographystyle{IEEEtran}
\bibliography{IEEEabrv, Retargeting_Running_CLF_RL}

\end{document}

%% file: sections/introduction.tex
\section{Introduction}
There has been immense progress in achieving dynamic motions on humanoid robots, but questions about how these motions can be generalized and controlled for autonomy still linger. Designing controllers that generate such behaviors is difficult as they require the ability to be both dynamic and robust. Reinforcement learning (RL) has emerged as a common tool of choice to synthesize such controllers due to its ability to leverage the full nonlinear dynamics, automatically adjust the contact with the environment, and domain randomize for robustness. In this work, we examine how reference tracking RL can be leveraged to generate dynamic and controllable humanoid running for application in an autonomy stack.

Bipedal and humanoid running has an extensive research history. Early work involved the Raibert heuristic \cite{raibert_legged_2000} for foot step placement in dynamic maneuvers, while later the the hybrid zero dynamics (HZD) methodology \cite{westervelt_hybrid_2003} was used to achieve running on planar bipedal robots \cite{ma_bipedal_2017, sreenath_embedding_2013, morris_achieving_2006}. HZD has also been extended to leverage human data \cite{ames_human-inspired_2014}, mirroring what modern controllers do. These methods used classical control concepts such as feedback linearization or control Lyapunov functions (CLFs) to stabilize the nonlinear system around reference trajectories that were generated through full order dynamic trajectory optimization. 

More recently, RL has been used for bipedal locomotion, including running. A bipedal robot completed the 100 m dash \cite{crowley_optimizing_2023} and separately RL was used for more versatile control which included running and the 400 m dash \cite{li_reinforcement_2024}. Yet, designing rewards for RL can be very difficult as a poor choice of reward can hurt the performance or training time. Therefore, a common approach to designing RL controllers for bipeds and humanoids is to reward the controller for tracking a pre-defined reference motion. This reference tracking RL paradigm can be done with diverse data sources such as a reduced order model (RoM) \cite{green_learning_2021}, full order dynamic trajectories \cite{li_reinforcement_2021}, and using human data \cite{xie_kungfubot_2025}. These ``mimic" style results, e.g. \cite{peng_deepmimic_2018, sleiman_zest_2026}, which track agile trajectories from human data, have shown exceptional agility but are generally restricted to single motion play back, while we are interested in generating steady state motions and using these in a controlled manner for autonomous locomotion.

\begin{figure}
    \centering
    \includegraphics[width=1.0\linewidth]{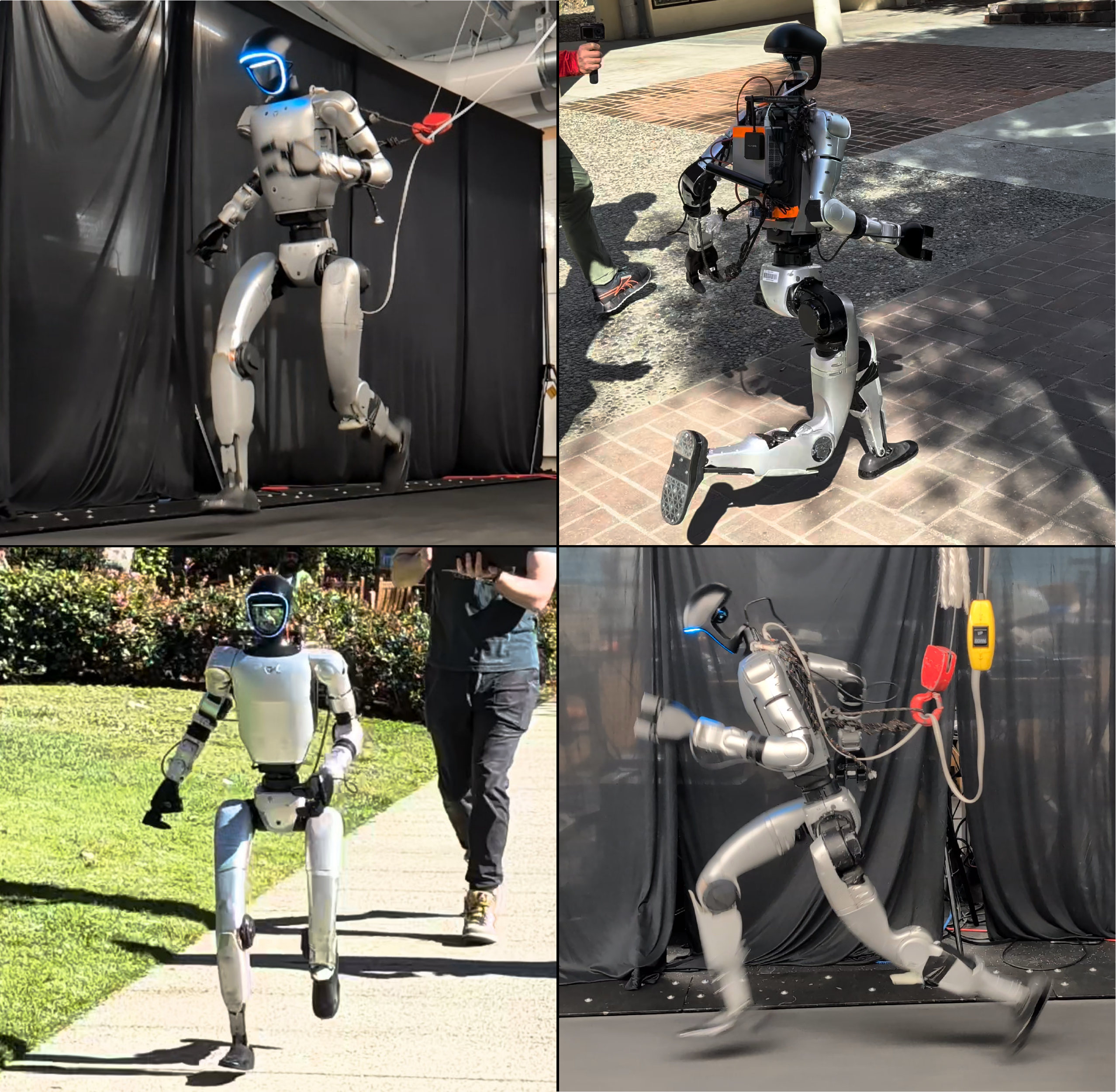}
    \caption{Demonstration of the running controller working inside on a constrained treadmill and in outdoor real world environments. The running appears human-like while still achieving the commanded speed through a combination of optimized retargeted human data and control guided reward shaping.}
    \label{fig:hero}
    \vspace{-8mm}
\end{figure}

We aim to design agile locomotion controllers that can be used by a higher-level planner in an autonomy stack. Autonomous robots are generally designed with multiple layers wherein each layer provides commands the layer below. Typically, locomotion controllers take in velocity commands, allowing a higher level controller to interface with the robot as a single integrator. If the resulting locomotion policy doesn't track commanded velocities, it is difficult to use the controller in a layered architecture. Historically, model-based controllers fit this paradigm well as we can generate tracking certificates and use those in the higher level control \cite{csomay-shanklin_bezier_2025}. This motivates both accurate velocity tracking, and investigation into use of model-based control ideas in RL locomotion controllers.  

\subsection{Contributions}
In this paper we answer three main questions:
\begin{enumerate}[label=Q\arabic*)]
    \item How can we generate/design agile reference motions that yield the best reference tracking performance?
    \item How does the reward structure effect both the ability to track references and the ability to track desired velocities?
    \item Can we integrate dynamic motions, like running, into a layered autonomy stack?
\end{enumerate}

We answer those questions through the following contributions.
    (C1) We propose to generate reference trajectories for RL controllers by retargeting human data through multiple-shooting dynamic optimization with constraints. This method has two major benefits over other works: (1) by adding in dynamics information and state constraints such as periodicity these trajectories can be tracked with lower error than only kinematically retargeted human data and (2) by adding constraints we can modify the trajectories in a dynamically feasible and principled manner to create a library of references.  
    (C2) We study the effect of different rewards on the quality of the reference and commanded velocity tracking. Mimic style rewards are compared with control guided rewards, i.e. CLF-RL \cite{li_clf-rl_2026}, for reference tracking performance. Different methods of tracking commanded velocities are compared including heuristically adjusting the trajectories, goal conditioning, and the effect of using a library of references over a single trajectory with goal conditioning. We conclude that goal conditioned CLF-RL with a library of dynamic optimized human data trajectories yields better performance than the baselines. 
    (C3) We demonstrate how the controllability can be used for heading tracking, lateral position control, and autonomy through extensive hardware experiments. The robot is able to achieve a top speed of 3.3 m/s on the hardware and can traverse paths hundreds of meters in distance in the real world. The controller can also be placed in an autonomy stack for obstacle avoidance. Fig. \ref{fig:hero} shows a few snapshots of the running experiments.

\subsection{Related Works}
\subsubsection{Reference Tracking RL}
There are many variations of the reference tracking RL concept such as tracking a RoM \cite{batke_optimizing_2022, green_learning_2021}, using a RoM feedback controller \cite{lee_integrating_2024}, tracking animated motions \cite{grandia_design_2024}, using full order dynamics \cite{li_reinforcement_2021, liu_opt2skill_2025}, or embedding a controller and dynamic references into the reward \cite{li_clf-rl_2026}. The most dynamic motions achieved through reference tracking RL track human data kinematically re-targeted onto the robot \cite{liao_beyondmimic_2025, peng_deepmimic_2018, xie_kungfubot_2025, sleiman_zest_2026}. 

The agility of these methods is exceptional, but for the most part the methods are restricted to re-playing single human motion clips, e.g. dance routines or kung fu moves. A diffusion policy to generate new reference motions was proposed in \cite{liao_beyondmimic_2025} but the motions generated from the diffusion don't yet have the same agility as the human motions. On the other hand, in \cite{peng_deepmimic_2018, li_reinforcement_2024} single references for running motions are tracked, and ``goal conditioned" rewards are used to make the robot track the desired commands, but this has not demonstrated accurate hardware velocity tracking yet.

\subsubsection{Motion Re-targeting}
When tracking human motions, the data must be converted to a format that the RL controller can use and generally the data needs to be filtered and cleaned before it can be tracked. Methods such as \cite{xie_kungfubot_2025, zhang_hub_2025} filter the human data based on stability heuristics and additionally smooth ground contacts. Alternatively, \cite{he_asap_2025} cleans and filters the data by learning a separate controller to determine if the motion is feasible. 
By using dynamic trajectory optimization, we can clean up the data, ensure it is dynamically feasible, and even improve it without heuristics or separate trainings.
Another approach to motion re-targeting uses sampling-based optimization \cite{pan_spider_2026}. With sampling based methods, enforcing state constraints, such as periodicity, is prohibitively difficult. Furthermore, such methods have yet to demonstrate success in highly dynamic and agile motions, finding success instead in quasi-static manipulation or low-speed humanoid locomotion. 

On the other end of the spectrum, dynamic trajectory optimization without any human data has been used to generate locomotion references for RL \cite{li_clf-rl_2026, liu_opt2skill_2025, li_reinforcement_2021}. This approach allows the motion to be highly optimized for the robot. In general these methods have been restricted to less dynamic motions, such as walking, although \cite{olkin_chasing_2025} achieved running without human data, the result had low time to overheat and showed minimal outdoor trials.
These methods can generate dynamically feasible and periodic motions, but they are not necessarily as well behaved as optimized human-data retargeted motions. 

\subsection{Outline}
\begin{figure*}
\vspace{2mm}
    \centering
    \includegraphics[width=1.0\linewidth]{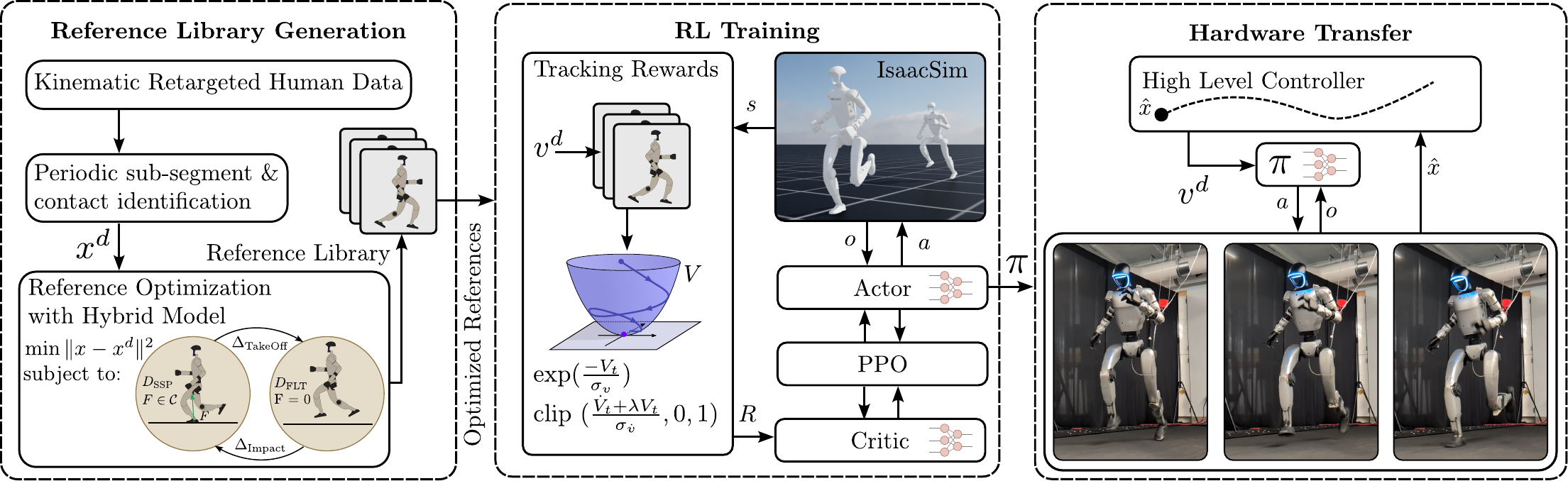}
    \caption{Architecture of the proposed pipeline. Human data is retargeted using state constrained dynamic optimization with a hybrid system model. This allows us to build a library of references from a single human reference and then track this in the RL controller. The RL controller uses a control-guided cost function. Then the policy is transferred zero-shot to hardware and can be used in an autonomy stack.}
    \label{fig:running_architecture}
    \vspace{-6mm}
\end{figure*}

The overall architecture of the method is shown in Fig. \ref{fig:running_architecture}. A library of dynamic reference trajectories are generated from a single human data segment. This reference is tracked in an RL training loop where a controller is synthesized. The resulting controller can then be applied on hardware with either joystick tracking or a layered control architecture to achieve controlled dynamic running.  
Section \ref{sec:prelim} provides a background on the hybrid systems model used for the optimization as well as a brief overview of CLFs. Then Section \ref{sec:reference_gen} presents the optimization-based retargeting to generate libraries of dynamically feasible trajectories. The different reward structures are presented in Section \ref{sec:reward} along with the overall RL environment design. Finally, in Section \ref{sec:results}, multiple simulation ablation studies are conducted and the proposed controller is applied to a real-world robotic humanoid (a Unitree G1), used outdoors over hundreds of meters, and in an autonomy stack.

%% file: sections/preliminaries.tex
\section{Preliminaries}
\label{sec:prelim}
\subsection{Hybrid Systems}
To formulate the dynamic retargeting problem, we use an explicit hybrid system model.
Legged locomotion can be modeled as a hybrid dynamical system with impulse effects, including both continuous dynamics and discrete events. We denote the system: $\mathscr{H} = (\mathcal{D}, \mathcal{S}, \Gamma, \Delta, \mathcal{F})$ where the domain $\mathcal{D} := \mathcal{X} \times \mathcal{U}$ consists of the state manifold and input space, the set of guards $\mathcal{S}$, the set of transitions $\Gamma$, the reset-maps $\Delta$, and the continuous dynamics $\mathcal{F}$. 
A transition from domain $D_i$ to $D_{i+1}$ is denoted as $\gamma := (D_i, D_{i+1}) \in \Gamma$. Then let $\mathcal{S}_{\gamma} \subset \mathcal{X}$ and $\Delta_\gamma$ denote the guard and associated reset map for a given transition, and $(f_{D}, g_{D}) \in \mathcal{F}$ denote the dynamics for a given domain $D \subset \mathcal{D}$.

We can write the hybrid dynamics of the system as 
\begin{numcases}{\mathcal{H} :=}
\dot{x} = f_{D}(x) + g_{D}(x)\,u_{D} & $x \in {D} \setminus \mathcal{S}_\gamma$, \label{eq: continuous_dynamics}
\\
x^+ = \Delta_{\gamma}(x^-) & $x^- \in S_\gamma$, \label{eq: discretecontrol}
\end{numcases}
where $x \in \mathcal{X}$, $u \in \mathcal{U}$.
The continuous dynamics can be represented:
\begin{align}
  M(q)\ddot{q} + H(q,\dot{q}) = B u + J_h(q)^TF \label{eq:dynamics} \\
  J_h(q) \ddot{q} + \dot{J}_h(q,\dot{q})\dot{q} = 0 \label{eq:hol_dynamics}
\end{align}
where $M(q):\mathcal{Q}\to \mathbb{R}^{n\times n}$ is the mass-inertia matrix, $H: \mathcal{Q} \times T\mathcal{Q} \to \mathbb{R}^n$ the Coriolis and gravity terms, and $B\in\mathbb{R}^{n\times m}$ is the actuation matrix. Contact constraints are enforced via the wrench $F \in \mathbb{R}^h$ and the constraint Jacobian $J_h(q) \in \mathbb{R}^{h \times n}$. Finally, the state is $x = [q, \dot{q}]^T$. See \cite{grizzle_3d_2010} for biped hybrid systems modeling details.

We consider two domains: the single support phase (SSP) and the flight phase (FLT) with transitions $(\text{SSP}, \text{FLT})$ (take off) and $(\text{FLT}, \text{SSP})$ (impact). These have the associated reset maps:
\begin{equation}
    \Delta_{(\text{SSP}, \text{FLT})}(x) = Ix.
\end{equation}
and
\begin{equation}
    \Delta_{(\text{FLT}, \text{SSP})}(x) : \begin{bmatrix} q \\ \dot{q} \end{bmatrix} \rightarrow \begin{bmatrix}
        q \\ 
        (I - M^{-1}J_h^T(J_hM^{-1}J_h^T)^{-1}J_h)\dot{q}
    \end{bmatrix} 
    \label{eq:impact_map} 
\end{equation}
with $I$ the identity matrix.

The domains are defined by the robot’s contacts with the ground, while the guards are determined by the ground height. Each domain specifies the number of active contact forces and their associated Jacobians. 

\subsection{Control Lyapunov Functions}
Control Lyapunov Functions (CLFs) are a classic tool in nonlinear control theory for designing certifiably stable controllers. Here we will be using the concept in the reward of an RL training routine.
For a smooth nonlinear control-affine system of the form \eqref{eq: continuous_dynamics}, a continuously differentiable, positive definite function $V(x) : \mathcal{X} \rightarrow \mathbb{R}$ is called an exponentially stabilizing CLF if the set
\begin{align}
    \{u \: | \nabla_x V(x)(f(x) + g(x)u)  < -\lambda V(x)\}
    \label{eq:clf_condition}
\end{align}
is non-empty for all states $x$ in an open region around the origin, for scalar $\lambda > 0$. This inequality guarantees that $V(x)$ decreases along system trajectories, implying exponential convergence of the state $x$ to the origin.

%% file: sections/methods.tex
\section{Reference Generation through Optimized Dynamic Retargeting}

\label{sec:reference_gen}

Although using human data directly is relatively easy because there is no need for a dynamics or contact model, there are a few issues with this approach: (a) there is still an embodiment gap between humans and humanoids, so a dynamically feasible motion on a human does not imply a feasible motion for a robot, (b) depending on the pipeline, the human data can be noisy, which then requires filtering before training, and (c) for steady-state locomotion, periodicity is required for the locomotion to be kinematically feasible, which in general is impossible to attain directly from human data. Therefore we introduce the use of a multiple shooting optimization problem to dynamically retarget the data, enforce periodicity, generate libraries of references, and otherwise clean the trajectory.

\begin{figure} 
    \vspace{2mm}
    \centering
    \includegraphics[width=1.0\linewidth]{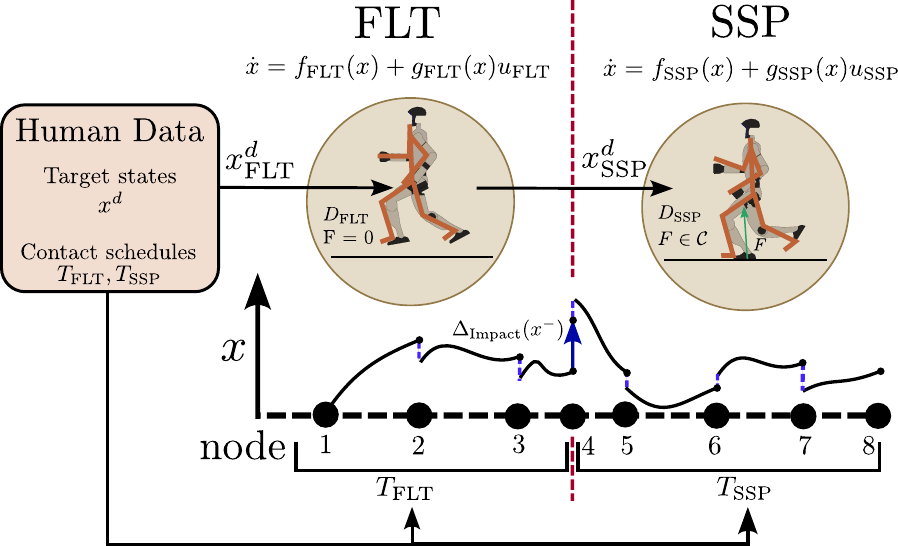}
    \caption{Visual depiction of the multiple shooting trajectory optimization problem. Two hybrid domains are shown and the node where they meet utilizes the associated reset map. Within each domain, a number of optimization nodes are used. The cost is tracking the human reference motion.}
    \vspace{-6mm}
    \label{fig:hybrid_opt}
\end{figure}

To start, we assume that we have some human data that has been retargeted onto the robots kinematics. Then, we formulate an optimization problem whose cost encourages the robot to mimic this human data while respecting hard constraints. Ultimately, a running gait library with speeds from 1.2 m/s to 3.6 m/s is generated from single piece of human data from the LAFAN dataset \cite{harvey_robust_2020}. To choose a portion of the dataset to use, we found a segment in which the human ran fast then we programatically extracted a single stride (both a single support and a flight phase) that was most periodic. To determine how periodic the reference trajectory is we symmetrically mirrored the joints, body positions, and orientations across the robot at the end of the stride and took the difference of this re-mapping compared to the original state. We only require a single stride as the other stride is always a mirror copy when traveling at steady state in a straight line. Fig. \ref{fig:hybrid_opt} gives a visual overview of the optimization process for enhancing human data.

To optimize the data, the problem is formulated with multiple shooting nodes \cite{diehl_fast_2006} which allows us to solve the optimization easier for unstable systems (i.e. a humanoid) compared to single shooting methods. We can also enforce hard state constraints, while other tools, like sampling based optimization, would struggle to do this. Solving a multiple shooting optimization problem for a hybrid system is difficult as to achieve global optimality the discrete decisions (the modes of the hybrid model) and the continuous actions should be chosen together. Solving this mixed-integer problem is computationally intensive and difficult to do. Therefore we choose to fix the hybrid domain sequence. We model running with two hybrid domains: the single support phase (SSP) and the flight phase (FLT). The time and order of the domains is extracted from the human data by thresholding the height of the foot above the ground. When in single support phase, a holonomic constraint and friction cone are used to enforce no foot slippage while in flight all ground reaction forces are set to 0. To encode a periodic steady state gait, the impact equations \eqref{eq:impact_map} are enforced on the reflected state. Therefore only a single stride is optimized and the other stride is always a mirror image.

To enforce the domain sequence, we create a set of node indexes $\mathcal{J}$ at which the domain must switch. Associated with each of these indexes are a corresponding guard and reset map. We denote the set of these guards and reset maps as $S_\mathcal{J}$ and $\Delta_\mathcal{J}$. The optimization only operates on discretized dynamics, which we denote as $\bar{f}_D$ and $\bar{g}_D$. This yields the full optimization problem as follows:
\begin{mini!}[1]
{\alpha, x, u}{\Phi(x, u)}
{\label{eq:hybrid-opti}}{}
\addConstraint{x_{k + 1}}{= \bar{f}_{D_k}(x_k) + \bar{g}_{D_k}(x_k)u_k \label{eq:opti-dynamics}}{\quad k = 0...N}
\addConstraint{x_{k}}{\in S_{k}}{\quad k \in \mathcal{J}}
\addConstraint{x_{k + 1}}{= \Delta_{k}(x_{k - 1})}{\quad k \in \mathcal{J}}
\addConstraint{x_{\text{min}}}{\leq c_x(x_k) \leq x_{\text{max}}}{\quad k = 0...N}
\addConstraint{u_{\text{min}}}{\leq c_u(u_k) \leq u_{\text{max}}}{\quad k = 0...N}
\end{mini!}
where $\Phi = \sum_{k=0}^N \|x_k -x^d_k\|^2_W$ with $W$ a weighting matrix, $x^d_k$ the state (base position, orientation, joint angles, and all velocities) of the human reference, and $c_u$ and $c_x$ denote generic nonlinear input and state constraints.

To solve the optimization problem, Casadi \cite{andersson_casadi_2019}, Pinocchio \cite{carpentier_pinocchio_2019}, and IPOPT \cite{wachter_implementation_2006} are used. To fix a forward speed we can constrain a forward step length based on the total stride time. The creation of a gait library is efficient by using solutions from the faster gaits as a warm start for solves of slower gaits, and modifying the step length constraint while maintaining step time. This allows us to generate incrementally slower gaits, and prevents neighboring gaits from converging to drastically different solutions. Each gait in the library attempts to track the same human reference.

Using optimization allows for rapid iteration, usually requiring seconds or minutes to solve for a gait. To transfer the gait to RL for training we fit a set of Bezier curves on desired ``outputs", denotes as $y^d(t)$. These Bezier curves are what is tracked by the RL reward, as discussed next. We choose to track all the joints, the pelvis frame and each of the end effector frames. Each frame is an SE3 pose which is represented by a 3-vector for positions and a quaternion for orientations. Also, all the velocities of the joints and twists of the frames are tracked.

\section{Reward Structure and RL Design}
\label{sec:reward}
\subsection{Reward Structure}
We analyze two separate reference tracking rewards: a mimic style reward where the bodies, their velocities, and joints are tracked in separate terms \cite{sleiman_zest_2026, liao_beyondmimic_2025, xie_kungfubot_2025}, and a CLF-RL reward where there is a single Lyapunov reward and a decrescent reward. Table \ref{tab:reward_table} gives a list of all the rewards for each of the methods and their weights. Since all the mimic rewards have the same form (except the quaternion which uses a box minus operator) we only list the expression once. All of the numerical values can be found in our open source code\footnote{\url{https://github.com/Zolkin1/robot_rl/tree/main}}. Note that the regularization and contact rewards are used for every method and the goal conditioning is only used in certain ablations.

\subsubsection{CLF Rewards}
Motivated by control Lyapunov functions (CLFs), we embed a Lyapunov function and its decrescent conditioning into the reward to encourage Lyapunov stable behavior. A quadratic Lyapunov function is used as we treat each output as a fully actuated double integrator \cite{li_clf-rl_2026}. Therefore the Lyapunov function takes the form of $V = \eta^T P \eta$ where $P$ is the solution to the continuous time Ricatti equation and $\eta$ denotes the output tracking error:
\begin{equation}
    \eta(t) = 
    \begin{pmatrix}
    y^d(t) - y(q) \\
    \dot{y}^d(t) - \dot{y}(q, \dot{q})
    \end{pmatrix}
\end{equation}
where $y(q)$ is the function mapping the configuration to the measured output.

The stability condition on the CLF is also put into the reward, where behaviors are penalized for not satisfying the CLF decrescent condition, i.e. if $\dot{V}_t + \lambda V_t > 0$.

\begin{table}
\vspace{1.5mm}
\centering
\caption{Reward terms used for each of the different training types.}
\label{tab:reward_table}
\begin{tabular}{p{2.8cm}lp{1.3cm}}
\toprule
\textbf{Reward Term} & \textbf{Expression}  & \textbf{Weight} \\
\midrule
\textbf{Regularization} & & \\
Torque                     & $\|\tau\|^2$       & $-1\times10^{-5}$ \\
Action-rate          & $\|a_t - a_{t-1}\|^2$ & $-1\times10^{-3}$ \\
Torque limit    & $\||\tau_t| - \tau_{\text{lim}} \|_1^+$ & $-1.0$ \\
Joint limit  & $\|(\underline{q} - q_t)^+ + (q_t - \bar{q})^+ \|_1$ & $-1.0$\\
\hline
\textbf{CLF-RL} & & \\
CLF tracking reward  & $\exp\left(\frac{-V_t}{\sigma_{v}}\right)$ & $10.0$ \\
CLF decay penalty & $\operatorname{clip}\!\left(\tfrac{\dot{V}_t + \lambda V_t} {\sigma_{\dot{v}}},\,0,\,1\right)$ & $-5.0$ \\
\hline
\textbf{Mimic} & & \\
Generic tracking (10x) & $\exp(-\|z - z^d\|^2 / \sigma_z)$ & 1.5 \\
\hline
\textbf{Contact} & & \\
Holonomic pos. & $\exp\left(-\frac{\|p_{\mathrm{st}} - p_{\mathrm{st}}^0\|}{\sigma_p}\right)$ & $4.0$ \\
Holonomic vel. & $\exp\left(-\frac{\|v_{\mathrm{st}}\|}{\sigma_v}\right)$ & $2.0$ \\
\hline
\textbf{Goal Conditioning} & & \\
Linear $(x,y)$ velocity & $\exp(-\| v - v^d \|^2/\sigma_v)$ & 1.0 \\
Yaw rate & $\exp(-\|\dot{\phi} - \dot{\phi}^d \|/\sigma_{\dot{\phi}})$ & 1.0 \\
\bottomrule
\end{tabular}
\vspace{-6mm}
\end{table}

\subsubsection{Mimic Rewards}
Mimic style rewards incentive tracking the reference trajectory too, but in multiple separate rewards and without the decrescent condition or the $P$ weighting matrix. Following common practice we choose all the weights to be the same and to track each group with a separate reward. The tracking groups include the joints, base position, base orientation, and end effector positions and orientations as well as all their linear and angular velocities, totaling to 10. The weight on the mimic rewards is set so that the total weight matches the total weight on the CLF rewards for a fair comparison. To tune the mimic rewards we started with the values given in \cite{sleiman_zest_2026} then tuned the standard deviations for performance in this application.

\subsubsection{Goal Conditioning and Velocity Tracking}
The mimic and CLF rewards are generally used only for reference tracking. Although the reference travels at a given velocity, it only moves straight, so the robot will not yaw or move laterally. There are two approaches that can be used at this point in the pipeline: (1) we can kinematically adjust the reference trajectory in the training loop to obey these motions or (2) we can add a goal conditioning reward. By kinematically adjusting the reference trajectory no additional reward is needed and therefore no additional tuning is required. Yet this is potentially a much weaker signal for the learning than goal conditioning. The goal conditioned reward incentivizes the robot's root velocity to match the commanded velocity, even if this contradicts the reference. 

\input{sections/tracking_table}

\subsection{RL Design and Training}
We use IsaacLab and IsaacSim for GPU-accelerated physics simulation \cite{nvidia_isaac_2025} and the PPO implementation from RSL-RL~\cite{rudin_learning_2022}. We use asymmetric actor critic where both the actor and critic networks use a fully connected feedforward architecture with three hidden layers, using the ELU activations at each layer. The actor's observations include the base angular velocity, projected gravity, commanded velocity, joint angles, joint velocities, the previous action, and $\sin$ and $\cos$ phases as observations. The phase observations have a period of exactly one full gait period. Note that the reference motion is not an observation to the policy which makes deployment very easy. To facilitate learning, the critic receives additional privileged information, such as end effector and base frame linear and angular velocities, reference trajectory positions and velocities, and binary contact state indicators.

The resulting policy should be able to track both joystick velocity commands and commands from a high level controller. To mitigate distribution shift with the commanded velocity, the policy is trained with three main groups of velocity commands: open loop velocity commands, heading control, and a P controller on both the heading and the lateral position. In this way the robot learns to obey the velocity commands while also being able to interface with a higher level controller. At episode reset the reference trajectory matching the desired speed is used for initialization, similar to what is done in \cite{peng_deepmimic_2018}.
Since it never starts standing, we instead choose to have 40\% of the environments start on a configuration from the fastest walking reference (1 m/s), encouraging the policy to learn the transition from walking to running. The same transition is learned for the walking policy to facilitate the run to walk transition. The running policy is trained on speeds between 1.1 and 3.7 m/s. During an episode, the commanded forward speed is used to index into the reference library.

Friction, mass, and motor properties are domain randomized. The full list and values can be found in our open source code. Additionally, a velocity-based push is periodically applied to the root of the robot. We find that sampling friction and gain randomization is very helpful for transferring the high speed running policy to hardware. The resulting control policy outputs joint angle setpoints at 50 Hz which are then tracked by high rate PD controllers. We use PD gains from \cite{liao_beyondmimic_2025} which over-damps the controllers. We find that this tends to prevent high frequency chatter on hardware.


%% file: sections/tracking_table.tex
\definecolor{winner}{RGB}{173, 216, 230}  

\begin{table*}
\vspace{0.5mm}
\centering
\caption{Velocity tracking error of the moving average (mean $\pm$ std) across policies and speeds. 
Bold indicates best per-metric value. \colorbox{winner}{Shading} indicates policy with lowest total error per speed.}
\label{tab:velocity_tracking}
\resizebox{\linewidth}{!}{%
\begin{tabular}{ll
    S[table-format=1.3] @{${}\pm{}$} S[table-format=1.3]
    S[table-format=1.3] @{${}\pm{}$} S[table-format=1.3]
    S[table-format=1.3] @{${}\pm{}$} S[table-format=1.3]
    S[table-format=1.3] @{${}\pm{}$} S[table-format=1.3]
    S[table-format=1.3] @{${}\pm{}$} S[table-format=1.3]
    S[table-format=1.3] @{${}\pm{}$} S[table-format=1.3]
    S[table-format=1.3] @{${}\pm{}$} S[table-format=1.3]
    S[table-format=1.3] @{${}\pm{}$} S[table-format=1.3]
    S[table-format=1.3] @{${}\pm{}$} S[table-format=1.3]
}
\toprule

& & \multicolumn{10}{c}{\textbf{CLF-RL}}
  & \multicolumn{8}{c}{\textbf{Mimic}} \\

\cmidrule(lr){3-12} \cmidrule(lr){13-20}

& & \multicolumn{4}{c}{Kinematic Retargeted Human Data}
  & \multicolumn{6}{c}{Dynamic Opt. Human Data}
  & \multicolumn{4}{c}{Dynamic Opt. Human Data}
  & \multicolumn{4}{c}{Kinematic Retargeted Human Data} \\

\cmidrule(lr){3-6} \cmidrule(lr){7-12} \cmidrule(lr){13-16} \cmidrule(lr){17-20}

Speed & Metric
  & \multicolumn{2}{c}{Goal + Adjust}
  & \multicolumn{2}{c}{3x Goal + Adjust}
  & \multicolumn{2}{c}{Adjust}
  & \multicolumn{2}{c}{Goal + Adjust}
  & \multicolumn{2}{c}{Goal Only}
  & \multicolumn{2}{c}{Goal + Adjust}
  & \multicolumn{2}{c}{3x Goal + Adjust}
  & \multicolumn{2}{c}{Goal + Adjust}
  & \multicolumn{2}{c}{3x Goal + Adjust} \\



\midrule

\multirow{3}{*}{1.4 m/s}
  & $e_{v_x}$ (m/s)
    & 0.437 & 0.054
    & 0.228 & 0.063
    & 0.251 & 0.106
    & 0.157 & 0.056
    & \cellcolor{winner} \textbf{0.139} & \cellcolor{winner} 0.045
    & 0.293 & 0.057
    & 0.236 & 0.050
    & 2.281 & 0.032
    & 0.438 & 0.110
    \\
  & $e_{v_y}$ (m/s)
    & 0.223 & 0.321
    & 0.196 & 0.322
    & 0.319 & 0.297
    & 0.221 & 0.313
    & \cellcolor{winner}\textbf{0.198} & \cellcolor{winner} 0.285
    & 0.221 & 0.310
    & 0.214 & 0.310
    & 0.264 & 0.274
    & 0.201 & 0.325
    \\
  & $e_{\dot\psi}$ (rad/s)
    & 0.214 & 0.241
    & 0.188 & 0.248
    & 0.202 & 0.229
    & 0.192 & 0.249
    & \cellcolor{winner} 0.216 & \cellcolor{winner} 0.258
    & 0.249 & 0.264
    & \textbf{0.187} & 0.235
    & 0.259 & 0.295
    & 0.200 & 0.248
    \\

\midrule

\multirow{3}{*}{2.0 m/s}
  & $e_{v_x}$ (m/s)
    & 0.335 & 0.083
    & 0.237 & 0.062
    & \textbf{0.096} & 0.081
    & 0.129 & 0.064
    & \cellcolor{winner} 0.104 & \cellcolor{winner} 0.058
    & 0.240 & 0.065
    & 0.206 & 0.055
    & 1.672 & 0.033
    & 0.599 & 0.128
    \\
  & $e_{v_y}$ (m/s)
    & 0.201 & 0.311
    & \textbf{0.181} & 0.309
    & 0.334 & 0.289
    & 0.209 & 0.310
    & \cellcolor{winner} 0.188 & \cellcolor{winner} 0.286
    & 0.213 & 0.299
    & 0.204 & 0.299
    & 0.252 & 0.273
    & 0.194 & 0.305
    \\
  & $e_{\dot\psi}$ (rad/s)
    & 0.230 & 0.254
    & \textbf{0.201} & 0.244
    & 0.207 & 0.234
    & 0.219 & 0.256
    & \cellcolor{winner} 0.232 & \cellcolor{winner} 0.264
    & 0.257 & 0.273
    & 0.205 & 0.234
    & 0.265 & 0.293
    & 0.248 & 0.263
    \\

\midrule

\multirow{3}{*}{2.6 m/s}
  & $e_{v_x}$ (m/s)
    & 0.298 & 0.077
    & 0.253 & 0.060
    & 0.316 & 0.130
    & 0.107 & 0.081
    & \cellcolor{winner} \textbf{0.081} & \cellcolor{winner} 0.081
    & 0.177 & 0.071
    & 0.162 & 0.058
    & 1.065 & 0.034
    & 0.605 & 0.121
    \\
  & $e_{v_y}$ (m/s)
    & 0.185 & 0.303
    & 0.178 & 0.298
    & 0.341 & 0.288
    & 0.192 & 0.299
    & \cellcolor{winner}\textbf{0.170} & \cellcolor{winner} 0.281
    & 0.205 & 0.289
    & 0.185 & 0.292
    & 0.246 & 0.278
    & 0.173 & 0.297
    \\
  & $e_{\dot\psi}$ (rad/s)
    & 0.248 & 0.275
    & 0.217 & 0.255
    & \textbf{0.214} & 0.239
    & 0.238 & 0.272
    & \cellcolor{winner} 0.245 & \cellcolor{winner} 0.279
    & 0.262 & 0.284
    & 0.223 & 0.250
    & 0.268 & 0.292
    & 0.266 & 0.282
    \\

\midrule

\multirow{3}{*}{3.2 m/s}
  & $e_{v_x}$ (m/s)
    & 0.246 & 0.081
    & 0.251 & 0.066
    & 0.674 & 0.136
    & 0.083 & 0.111
    & \cellcolor{winner}0.076 & \cellcolor{winner} 0.123
    & \textbf{0.070} & 0.113
    & 0.090 & 0.088
    & 0.456 & 0.045
    & 0.484 & 0.103
    \\
  & $e_{v_y}$ (m/s)
    & 0.171 & 0.292
    & 0.172 & 0.286
    & 0.342 & 0.290
    & 0.173 & 0.289
    & \cellcolor{winner}\textbf{0.157} & \cellcolor{winner} 0.278
    & 0.197 & 0.286
    & 0.177 & 0.287
    & 0.239 & 0.287
    & \textbf{0.157} & 0.290
    \\
  & $e_{\dot\psi}$ (rad/s)
    & 0.262 & 0.292
    & 0.232 & 0.262
    & \textbf{0.222} & 0.248
    & 0.252 & 0.289
    & \cellcolor{winner} 0.255 & \cellcolor{winner} 0.291
    & 0.273 & 0.292
    & 0.235 & 0.261
    & 0.271 & 0.296
    & 0.272 & 0.285
    \\

\bottomrule
\end{tabular}
}
\vspace{-4.5mm}
\end{table*}

%% file: sections/results.tex
\section{Results}
\begin{figure}
    \centering
    \includegraphics[width=1.0\linewidth]{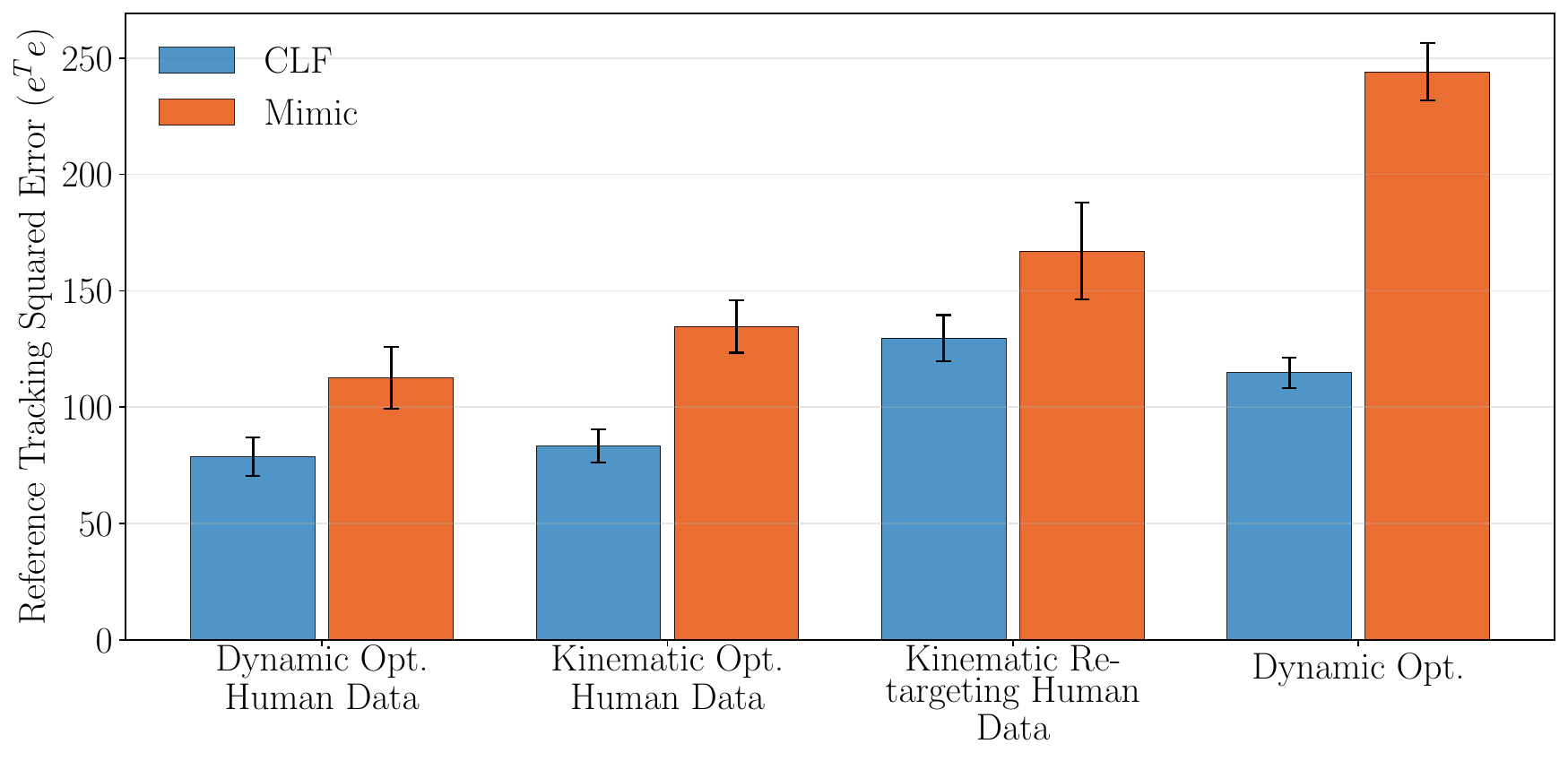}
    \caption{Reference tracking comparison between reference and rewards types. Error bars show one std. dev. The dynamically optimized human data performs the best, and the CLF rewards outperform the Mimic rewards. Optimized retargeted references outperform kinematic human data retargeting.}
    \label{fig:ref_rew_abl_bar_chart}
    \vspace{-6mm}
\end{figure}

To evaluate the effect of the optimized reference motions and the different reward structures we perform two simulation experiments. Then we apply the controller to the real-world robot for indoor and outdoor testing.
\begin{figure*}
\vspace{1.1mm}
    \centering
    \includegraphics[width=1.0\linewidth]{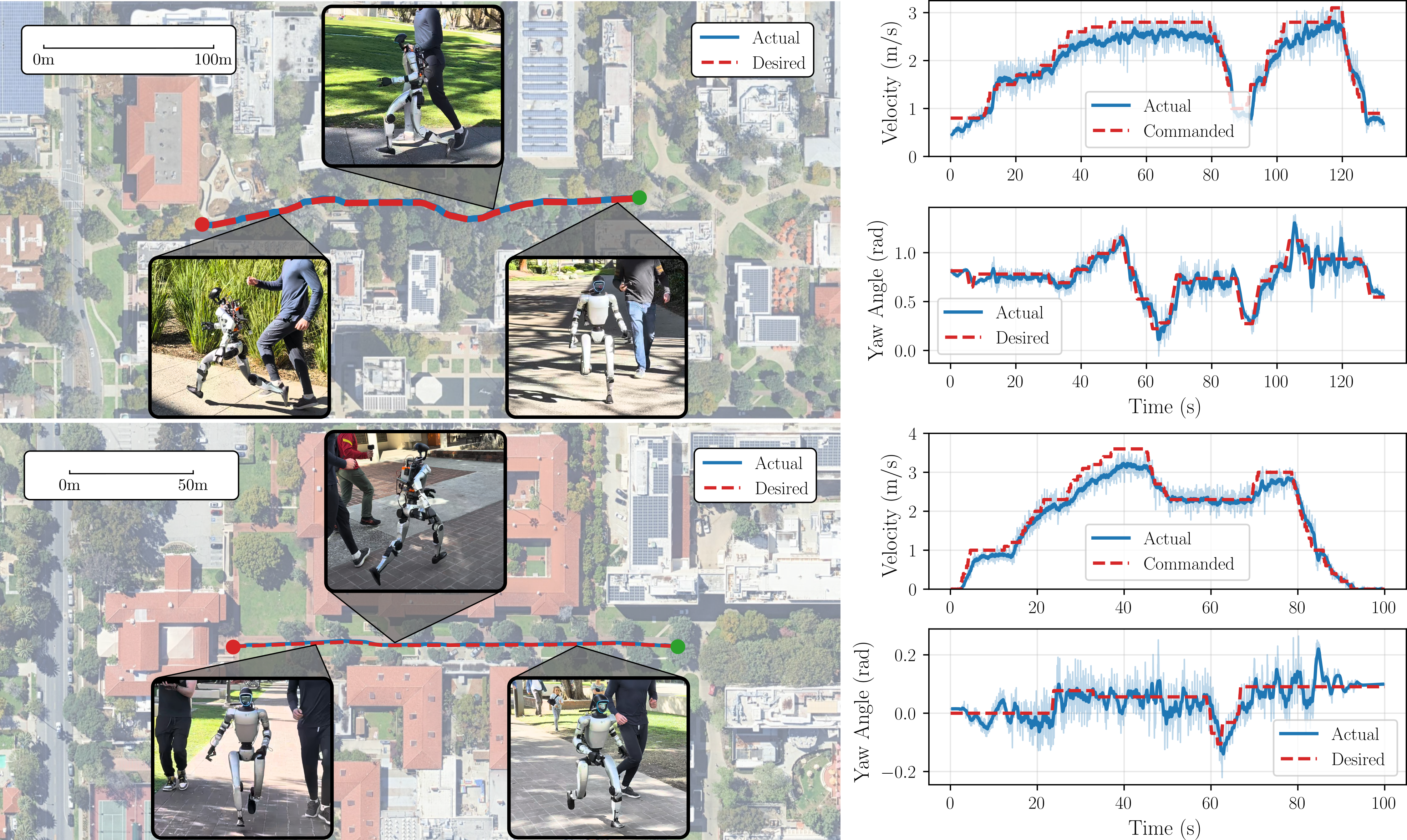}
    \caption{Outdoor long range experiments. The robot runs outdoors over distances of more than 150 m. The robot was given yaw targets in the global frame, a target position to track with body lateral motion and a feedforward velocity in the body forward direction. Lidar odometry and a P controller were used for global pose tracking.}
    \label{fig:global_pose_tracking}
    \vspace{-6mm}
\end{figure*}

\label{sec:results}
\subsection{Reference Motion and Tracking Reward Ablation Study}
We compare the effects of using optimization to dynamically retarget the reference motion vs both generating a reference motion without any human data and tracking the kinematically retargeted trajectory. We also ablate the effect of using the optimization, but without any dynamics (i.e. kinematic optimization). The RL policies are trained to track a single reference motion which allows us to examine the reference tracking alone without worrying about velocity tracking or libraries of references. There are two groups of policies trained here: one group using the CLF-RL rewards and the other using the mimic rewards. Within each group there is a policy trained on each of the 4 types of references. Other than the reward differences between groups and the reference motions between policies, the environments of the training are identical. It should be noted that for the dynamic optimization without human data that the optimizer would not converge at speeds over 3 m/s while all the human data based references are at 3.6 m/s.

Once the policies are trained we generate the experiment data by spawning 200 robots in IsaacSim with domain randomization. We then play the simulation for 7 seconds giving the robots the single velocity command they were trained on. The reported error metric is the mean over the environments of the squared tracking error averaged across time. Fig. \ref{fig:ref_rew_abl_bar_chart} shows the collected simulation data. There are two main takeaways from this plot: (1) Across both rewards we can see that optimizing the human data results in more accurate reference tracking than kinematic retargeting, and dynamic optimization out-performs kinematic optimization, (2) for every reference type the CLF reward yields a lower tracking error than the mimic rewards. This answers our question (Q1) by demonstrating that optimized human data out-performs kinematically retargeted human data for reference tracking. We can also see that the optimized human data out-performs a hand-crafted dynamic optimization that doesn't have human data. These results imply that the quality of the trajectory is critical for accurate tracking.

This plot also answers part of our question (Q2) regarding the effect of reward structure on tracking by demonstrating that CLF-RL rewards out-perform the mimic rewards. This highlights the fact that reward structure matters. In the case of CLF-RL we design the reward to directly encourage theoretically sound stability to the reference.

\subsection{Velocity Tracking Ablation Study}
Now we can study which reward and reference combination yields the best velocity tracking which is critical to embedding dynamic motions in an autonomy stack. Velocity tracking performance is tested in Mujoco after the policies are trained in IsaacLab to help verify sim-to-sim performance. These policies are all trained to track a variety of speeds: $v_x \in [1.1, 3.7], \; v_y \in [-0.75, 0.,75], \; v_{\dot{\phi}} \in [-0.75, 0.75]$. Here we ablate the reference type between dynamic optimized human data and kinematically retargeted human data and we also ablate the tracking reward between mimic and CLF-RL, then finally we examine the effect of kinematic trajectory adjustments, goal conditioning on a library of references and goal conditioning on a single kinematically retargeted human trajectory. We choose to compare the dynamic optimized human data because this performed best in our first ablation study and we choose to compare to kinematically retargeted human data with goal conditioning because this is a baseline that has been done before \cite{li_reinforcement_2024, peng_deepmimic_2018}, and is a very natural next step for a mimic-based pipeline that can re-play human motions. This helps to address the question whether using the library generation from the retargeting optimization provides value.

We compare each of these policies tracking the given target speed (in the body forward direction) moving straight, as well as with the lateral and yaw rate inputs set to their max values: $v_y^d = \pm 0.75$ and $v_{\dot{\phi}} = \pm 0.75$ for different subsets of the simulation time. Domain randomization on the torso mass, CoM position, and D gains for each joint is applied in Mujoco and 50 trials at each speed are logged in Table \ref{tab:velocity_tracking}.

There are three main conclusions we can draw from the data: (1) although only kinematically adjusting the trajectory does achieve some amount of tracking, goal conditioning provides a better signal in the training process, (2) the CLF reward in general facilitates better velocity tracking, and (3) even across rewards the dynamic optimized reference trajectory yields better tracking than goal conditioning a single human trajectory, even if we increase the the goal conditioning weight by 3x. We increased the goal conditioning weights by 3x on the kinematic retargeting to address the fact that the reference motion is not consistent with the commanded velocity so there will inevitably need to be more deviation. This finishes the answer to our question (2): we can see that goal conditioning improves the velocity tracking but it is best paired with a dynamic reference library and a CLF tracking reward.

The dynamic optimization of the human data also immensely improves the tracking at lower speeds which is mostly due to the fact that the reference motion from the human is a high speed run. Therefore the ``style" that comes from that motion is most accurate for faster speeds. This is why in general the forward velocity error increases with slower speeds: because the reference is fast. The dynamic optimization allows us to get better tracking from a single human data trajectory for a much larger range of velocities.

\subsection{Hardware Results, Integration with Autonomy Stack}
We deploy the policy on a Unitree G1 robot on an indoor treadmill environment, in large scale outdoor tests, and in autonomous obstacle avoidance scenarios. The running includes a full flight phase with the robot completely in the air. All policies and computation are run on-board using a mini-pc powered by the robot. Running the robot on an indoor treadmill (see Figs \ref{fig:hero} and \ref{fig:running_architecture}) demonstrates that the robot achieves high speeds (up to 3.3m/s) in a tight environment. Additionally, the robot has run more than 250 m consecutively on outdoor sidewalks. Fig. \ref{fig:global_pose_tracking} shows two example runs. In these experiments a feedforward velocity is set by a remote and the robot tracks global yaw angles and tracks a global position target with only lateral motion using a P controller, which sends velocity commands to the running controller. In this way the running controller is operating in a layered control stack. We get position and yaw estimates from FastLio \cite{xu_fast-lio_2021} at 10 Hz. The velocity data in Fig. \ref{fig:global_pose_tracking} is obtained by finite differencing the odometry data.

\begin{figure}  
    \vspace{2.5mm}
    \centering
    \includegraphics[width=1.0\linewidth]{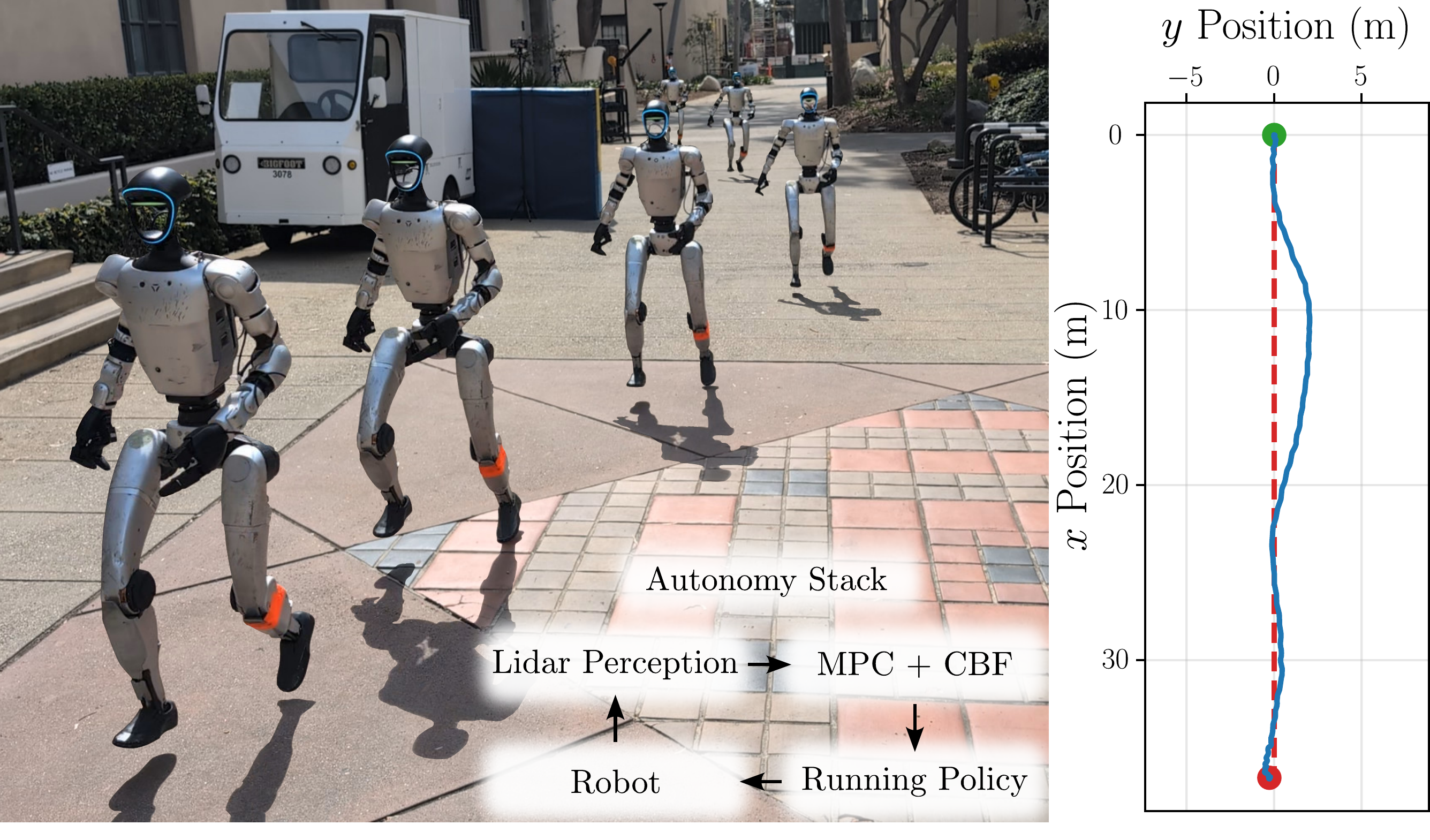}
    \caption{Autonomous collision avoidance while running with CBF-constrained MPC. The robot perceives it environment with a lidar, continuously re-computes a plan to avoid collisions, and passes the corresponding velocity commands to the running controller. With this architecture, the robot can avoid real-world obstacles while running.}
    \label{fig:cbf_avoidance}
    \vspace{-5mm}
\end{figure}

The running can also be controlled to perform collision avoidance within an autonomy stack. Specifically, we implement a model predictive control (MPC) algorithm that generates velocity commands for the running controller. This MPC layer uses real-time lidar scans to construct and update a local occupancy map, as in \cite{yamaguchi_layered_2026}, which is then used to synthesize a smooth Poisson safety function \cite{bahati2025dynamic}. We employ the Poisson safety function to formulate a control barrier function (CBF) collision avoidance constraint along the MPC horizon. The robot is commanded to run straight ahead while the MPC + CBF layer 1) characterizes the environment, 2) plans a new path, and 3) updates the running command to avoid collisions. The resultant controller is agile enough to maintain forward speeds around 2 m/s while dodging detected obstacles. Fig. \ref{fig:cbf_avoidance} shows a demonstration of this end-to-end collision avoidance capability in a real-world outdoor environment.

